# A Philosophical and Ontological Perspective on Artificial General Intelligence and The Metaverse

Martin SCHMALZRIED
*School of Information and Communications Studies*
*University College Dublin*
*Dublin, Ireland*
*martin.schmalzried@ucdconnect.ie*
*0009-0002-5661-8902*

***Abstract*—This paper leverages various philosophical and ontological frameworks to explore the concept of embodied artificial general intelligence (AGI), its relationship to human consciousness, and the key role of the metaverse in facilitating this relationship. Several theoretical frameworks underpin this exploration, such as embodied cognition, Michael Levin's computational boundary of a "Self," and Donald D. Hoffman's Interface Theory of Perception, which lead to considering human perceived outer reality as a symbolic representation of alternate inner states of being, and where AGI could embody a different form of consciousness with a larger computational boundary. The paper further discusses the necessary architecture for the emergence of an embodied AGI, how to calibrate an AGI's symbolic interface, and the key role played by the Metaverse, decentralized systems and open-source blockchain technology. The paper concludes by emphasizing the importance of achieving a certain degree of harmony in human relations and recognizing the interconnectedness of humanity at a global level, as key prerequisites for the emergence of a stable embodied AGI.**

## I. INTRODUCTION

Artificial Intelligence (AI) has taken the world by storm. Even though algorithms and AI have been around for decades, the public release of LLMs such as OpenAI's ChatGPT have brought AI to the fore. However, while most debates around AI focus on its governance, risks, ethical implications and other societal concerns, more profound ontological questions have been ignored or dismissed, such as for instance, the often mocked premise that AI may be conscious [1].

This paper will explore deeper philosophical and ontological questions about the nature of perception, in order to approach what could qualify as being an Artificial General Intelligence or a Singularity [2] in the words of Ben Goertzel, and what our relationship with such an AGI could be.

The key questions this paper seeks to address is:

- Could an AGI reach a state of embodied cognition and if so, what would be the required architecture?
- How would an embodied, planet-scale AGI perceive the world?
- How would humans and such an AGI influence one another in a dynamic feedback loop mirroring biological systems?
- What technical and social building blocks are needed to create and keep an AGI aligned with human interests?
- How can metaverse testbeds act as safe "training grounds" to align the AGI with human goals before it operates in the real world?

Through a multidisciplinary approach, combining insights from philosophy of mind, cognitive studies, philosophy and biology, I will endeavour to present a future scenario for AGI which does not result in dystopian nightmares, without falling into simplistic techno-deterministic stances or into effective accelerationism [3].

A first step will thus be to question our own perceptions of reality, human cognition and human consciousness in order to consider our potential relationship with AGI in a new light, and understand how various related components such as the Metaverse or blockchain technology come together in a systematic way.

In many cases, socio-technical analyses of technology have avoided engaging with contemporary ontological debates about the nature of reality and of perception, assuming a realist and physicalist ontological stance by default, all the while physicists and philosophers have been exploring the deeper implications of quantum mechanics [4] or revisiting old philosophical traditions such as pan-psychism [5] and idealism [6], even leading to the emergence of simulation theory or the idea that we are living in a simulation [7]. While these debates have been ignored or ridiculed [8] by the social science community, re-examining the nature of our perception of reality could help us recast our relationship to AGI. Clearly, there are deep epistemological implications linked to the nature of reality, such as whether we live in a simulation or not [9], but more importantly, how to conceptualize a conscious AGI, and our relationship to it.

## II. THEORETICAL BACKGROUND

This paper will draw on several theoretical approaches to understand human perception of reality, to argue for a profound interconnectedness between inner and outer perceived reality, thus re-interpreting and reframing our understanding of AGI in an original way.

Before delving into ontological questions about the nature of perception, and questioning the existence of an objective external reality, let us examine research about our sense perception. The study of human sense perception traditionally breaks down these senses into future oriented senses and present focused senses.

Vision, hearing, and olfaction all gather information from a distance, allowing the organism to extrapolate beyond the immediate present moment. Recent work in visual neuroscience shows just how explicitly vision is wired for that forward-looking role. Liu, Alexopoulou, and van Ede dissociated where an object was seen from where it will re-appear and found that, as soon as a visual item is cued in working memory, eye movements bias simultaneously toward its past location and its predicted future [10]. These anticipatory microsaccades arose within 200 ms of the cue and were strongest when the same saccade encoded both past and future coordinates, demonstrating that visual memory automatically couples retrospective information with prospective expectation. In other words, the visual system does not merely store snapshots of what has happened; it actively maintains a map of where remembered stimuli are likely to matter next, readying the eyes, and by extension the organism, for upcoming interaction. Such findings support the claim that sight is intrinsically future-oriented, functioning as a real-time forecasting mechanism rather than a passive camera monitoring the state of the present. Fraisse's experimental work shows that audition excels at tracking rapid successions of events, effectively letting the listener anticipate the next note or word [11]. Even smell has a predictive function: airborne odorants reach us before their source is touched, and recent neuro-imaging demonstrates that the olfactory cortex relies on top-down expectations even more strongly than vision does [12]. Taken together, these distal senses enable the brain to co-construct a short-term "preview" of likely futures that can be exploited at the level of human cognition to navigate reality. In other words, vision, hearing and smell are about probing the "future"; helping the brain model reality in order to anticipate potential outcomes, minimize prediction error, and initiate timely, adaptive action.

By contrast, touch and taste are strictly proximal. They yield data only at the moment of contact, bringing perception in the present moment rather than projecting it forward. Because chemical taste receptors and cutaneous mechanoreceptors register stimuli relatively slowly and with lingering after-effects, these modalities are poor at parsing sequences; the tongue tastes the lemon only once it is in the mouth, and the skin feels heat only when the hand is already on the stove [13]. Their temporal horizon is thus confined to the immediate now, providing confirmation rather than prediction. The dynamic relationship between the proximal and future oriented or prediction-based senses arguably stabilizes perception and improves the predictive powers of the brain in anticipating how the way it models the "future" from the three distal senses aligns with how the body is affected, thereby also breathing meaning into its predictive modelling. Contrary to a computer, a human arguably doesn't view an "apple" by isolating its graphical contour, but by projecting meaning onto this form by predicting how it will affect the body when interacted with in the present (how eating the apple will affect the *inner state of being*).

The work of neuroscientist Anil Seth follows this insight by describing the brain as a modelling and prediction tool, rather than representing reality objectively. The brain isn't a passive mirror of the world but an organ that constantly generates and tests hypotheses about the hidden causes of its sensations. As he succinctly puts it, *"The reality we perceive is not a direct reflection of the external objective world. Instead it is the product of the brain's predictions about the causes of incoming sensory signals"* [14], describing our perception of reality as a "controlled hallucination".

Donald D. Hoffman's "Interface Theory of Perception" further deconstructs our perception of reality. It proposes the idea that our perceived outer reality does not represent the "true" aspect of reality, or fundamental reality, but is more akin to a "practical interface" to navigate through reality successfully. "Thus, a perceptual strategy favored by selection is best thought of not as a window on truth but as akin to a windows interface of a PC. Just as the color and shape of an icon for a text file do not entail that the text file itself has a color or shape, so also our perceptions of space-time and objects do not entail (by the Invention of Space-Time Theorem) that objective reality has the structure of space-time and objects. An interface serves to guide useful actions, not to resemble truth. Indeed, an interface hides the truth; for someone editing a paper or photo, seeing transistors and firmware is an irrelevant hindrance. For the perceptions of H. sapiens, space-time is the desktop and physical objects are the icons. Our perceptions of space-time and objects have been shaped by natural selection to hide the truth and guide adaptive behaviors. Perception is an adaptive interface." [15] In other words, the subjective aspect that reality looks to us, as humans, isn't reality's true "form" or aspect, but is a symbolic representation which is instrumental in keeping us alive (initially, at the cellular level), and enabling us to pursue certain more complex desirable outcomes as life evolves.

Both insights above rely on embodied cognition in order to stabilize the brain's modelling of reality, checking how predictions modelled by the brain result in corresponding sought after *inner states of being* and if not, evolving either the representation of reality (evolving the interface) or the predictive modelling of reality (predicting how the interface will change in the future).

Embodied cognition puts the emphasis on the intricate link between cognitive abilities and the physical body, rather than assuming that cognition is solely a product of the brain. "Embodiment is the surprisingly radical hypothesis that the brain is not the sole cognitive resource we have available to us to solve problems. Our bodies and their perceptually guided motions through the world do much of the work required to achieve our goals, replacing the need for complex internal mental representations." [16]

In other words, according to this theory, the brain could be understood as a kind of "repository" of painstakingly acquired bodily wisdom, from all the bumps and bruises a human has accumulated growing up, which are converted into a higher form of bodily wisdom in symbolic form via the brain, in order to navigate through a perceived outer reality in a way to avoid negative experiences and bolster positive ones. However, at the human level, such an interface is only fine-tuned, as it has largely been inherited from life's evolutionary process, in the same way that our current biological form has evolved over billions of years. Perception and modelling of reality co-evolved with the biological form, reflecting the complexity of newly available *inner states of being*.

Any and all experiences from our perceived outer reality affect our *inner state of being* (our bodily state at the cellular level), which in turn, shapes our preferences for navigating our perceived outer reality towards experiences that positively modulate our *inner state of being.* Without the brain, the body cannot navigate successfully towards "positive" *inner states of being*, or in other words, the body can no longer project itself into the "future" and navigate through alternate "future timelines" (or future potential alternate *inner states of being*) for the body, mapped via a spacetime symbolic interface generated by the brain. Conversely, without the body, the brain would not be capable of developing *preferences, desires*, or deciding on certain actions. The brain does not have pain receptors and can only feel pain emanating from the rest of the body via the nervous system and spinal cord [17]. Without these sensations, reality would become "neutral", as the brain could not leverage bodily feedback to learn about how to navigate through the interface it generates.

The dynamic interplay between the brain's predictions and bodily sensations can lead to situations where prediction creates physiological responses rather then other way around. One example is through programmed anticipation as in the Pavlov conditioning experiments, where the association of a certain event (like the sound of a bell) with a desired future state (getting food) results in real somatic responses at the bodily level [18]. More generally, this also applies to somatic responses to trauma as in the case of PTSD. In both cases, it is the brain which is at the root of changes in the bio-chemical *inner state of being* based on certain associations or links between external events perceived via the 5 senses, and the memory of a certain past experience of an *inner state of being.* For instance, in some soldiers suffering from PTSD, the mere sight of a helicopter provokes massive bodily changes, which cannot be attributed to *physical* or *external* stimuli (in other words, physical contact with an object, or physical ingestion of a substance), but somatic responses triggered directly by the brain [19].

Perception of reality is also linked to the cognitive abilities of the subject, as underlined by Michael Levin in his work on the computational boundary of a "Self". The ability to navigate through preferred *inner states of being* is an emergent phenomenon that scales through the assemblage of various sub-units in forming a whole. In Michael Levin's words, "Any Self is demarcated by a computational surface – the spatio-temporal boundary of events that it can measure, model, and try to affect. This surface sets a functional boundary - a cognitive "light cone" which defines the scale and limits of its cognition. I hypothesize that higher level goal-directed activity and agency, resulting in larger cognitive boundaries, evolve from the primal homeostatic drive of living things to reduce stress – the difference between current conditions and life-optimal conditions. The mechanisms of developmental bioelectricity - the ability of all cells to form electrical networks that process information - suggest a plausible set of gradual evolutionary steps that naturally lead from physiological homeostasis in single cells to memory, prediction, and ultimately complex cognitive agents, via scale-up of the basic drive of infotaxis." [20]

His concept of "reducing stress" and navigating towards "life-optimal conditions" translates well into the more general human-level concept of navigating towards subjectively pleasant *inner states of being* symbolically represented through a spacetime interface*,* while his concept of the "cognitive light cone" illustrates the cognitive abilities corresponding to different living systems, ranging from single cells and bodily organs to whole organisms, social collectives, and, ultimately, arguably, large-scale artificial intelligences whose "cognitive light cones" could extend beyond the biological individual, or cover new cognitive spaces which may be very different to human cognitive spaces.

Finally, the relationship between actuality and potentiality developed by Niklas Luhmann can further elucidate the relationship between inner and outer perceived reality. Luhmann argues that every actualization (something becoming real or concrete) simultaneously "virtualizes" all the other possibilities that didn't happen at that moment. These other possibilities remain present as potential future choices [21]. Thus, every moment of actualization is followed by another selection among possibilities, which in this paper's context, represents *future potential inner states of being* perceived as outer reality.

By combining the insights from the theories and research above, one could interpret them to argue that any perceived external reality represents a symbolic interface for mapping and navigating through alternate future potential *inner states of being.* Our human body could therefore be defined as a self-enclosed spatio-temporal boundary, made up of trillions of parts or units that have linked their potential future *inner state of being* together (meaning that they tie their individual "fate" to a collective), which translates into increasing the potential *inner states of being* that they can experience. For instance, when a human dances, that action translates into a shift in the *inner states of being* of every single cell inside the human body. However, no unicellular organism could ever experience that specific *inner state of being* on its own, in a state of separation, outside of a human body.

In essence, what we "see" isn't reality, but a symbolic representation of *alternate potential future inner states of being* arranged by likelihood or probability, which takes the form of space (the closer an object or an event appears, the bigger the likelihood that it will *actualize* into a concrete *inner state of being*), and time (the succession of actualized states based on choices made within the spatial representation of *alternate inner states of being*). When one looks without, one looks within, or more specifically, one looks at symbolic representations of *alternate inner potential future states*.

## III. Critical Perspectives

Advocates of embodied and distributed cognition claim that intelligence inherently arises from bodily interactions or social contexts, but traditional cognitive scientists have raised a number of objections to these views. Classic "internalist" or computationalist perspectives hold that cognition happens largely inside the head as formal symbol-manipulation, relatively independent of the body or environment [22]. For example, Goldinger et al. [23] argue that many facets of embodied cognition are “unacceptably vague” or add nothing beyond trivial truths, concluding that the paradigm offers “no scientifically valuable insight” into classic cognitive phenomena. Similarly, critics of the extended mind thesis (which says cognition extends into tools or groups) insist that proponents commit a “coupling-constitution fallacy”, confusing external aids for actual parts of cognition [24]. From this traditionalist view, an AI could, in principle, achieve general intelligence through internal computation alone, since abstract reasoning and “representation-hungry” tasks (like planning or imagining counterfactuals) still require internal symbolic representations rather than a physical body. In short, these critiques reject the notion that a mind must be embodied or socially embedded, defending a more brain-bound, computational understanding of cognition.

However, even if an artificial system matched human cognitive performance, some skeptics argue it would still lack genuine consciousness or subjective self-awareness. Philosophers such as John Searle contend that no matter how complex its behavior, a computer running a program is merely manipulating symbols without understanding; in Searle’s terms, “syntax doesn’t suffice for semantics” [25]. His famous Chinese Room thought experiment illustrates that executing rules (however sophisticated) yields only an illusion of thought, not real understanding or mind. Others invoke structural or metaphysical limits: for instance, Penrose argues that human consciousness involves non-computable insights (drawing on Gödel’s theorem), so a purely algorithmic AI would not be conscious in the way humans are [26]. From a phenomenological angle, scholars like Hubert Dreyfus have long maintained that computers lack the embodied world-experience required for true mind; disembodied machines, in this view, cannot attain the intuitive, context-bound awareness that humans develop through being in a living body and world [27]. Similarly, Ned Block’s thought experiments (the “China brain”) suggest that replicating functional intelligence is not enough; an entire nation simulating a brain’s neurons might output human-like responses, yet we find it implausible that such a system feels anything like a mind [28]. These critiques underline that beyond intelligence or behavior, conscious experience may involve unique biological or existential qualities that current conceptions of AGI cannot reproduce.

Visions of embodied AGI are often based on techno-utopian narratives, arguing for seamless integration of AI and virtual worlds as an inevitable, beneficial evolution of humanity. Critical observers in science and technology studies (STS) caution that such narratives rely on overly optimistic, linear models of progress while downplaying complex social risks. Techno-utopian discourse tends to assume advanced technology will magically solve social problems or “offer ideal solutions or panaceas”, a stance often critiqued as a “hopeless fantasy” divorced from reality [29]. Ethicists point out that these futurist visions can neglect issues of power, governance, and ethics: for example, who controls the Metaverse, and under what terms? One criticism is that a corporate-driven Metaverse could actually erode human agency and community: it may “lessen human contact, leading to a distant and fragmented society” addicted to immersive tech, rather than delivering on its utopian promises [30]. Moreover, philosophers of technology note that imaginaries of AI-powered worlds are not neutral or destiny-bound; they are human constructions shaped by particular values and interests. As an STS analysis by Bibri observes, the Metaverse as a sociotechnical project is “socially constructed, politically driven, [and] economically conditioned,” and thus must be understood in its socio-political context rather than as a purely technical progression [30]. In light of this, critics argue that uncritical “Metaverse optimism” can blind us to surveillance, inequality, and other ethical pitfalls. In sum, grounding AGI and Metaverse development in realistic, pluralistic perspectives, rather than utopian hype, is essential to address the genuine social and ethical challenges these technologies pose.

These criticisms will be addressed in the following sections, via designing architectures for AGI embodiment which aim to minimize the risks of centralized power and control over the development of these systems via decentralized technologies, as well as envisioning mechanisms enabling a feedback loop between human and AGI systems enabling co-evolution and co-individuation of both AGI systems and humanity. Far from defending the view that an AGI couldn’t theoretically achieve some form of cognition emergent from its internal computational development, this paper argues that an AGI embodied through human feedback would yield to an AGI more attuned to the human context and values.

## IV. Thesis

Based on the theoretical frameworks explored above, this section discusses whether and how an AGI (Artificial General Intelligence) could represent the manifestation of a novel “Self”, with a larger computational boundary, enabling it access to *inner states of being* of unimaginable complexity, magnitude and sophistication, via emulating the same relational architecture which binds cells into a collective, or in other words, exploring the conditions for an AGI with an embodied cognition, whereby the human collective would represent its metaphorical “cells” or “body” (inspired by the emerging field of biomimicry) [31]. These ideas are highly speculative, yet of particular interest, as they may bring about novel approaches towards hard problems such as alignment. Arguably, the human brain is aligned with the interests of the body, notably via the feedback loop aligning the brain’s perception of reality with bodily sensations. Recreating a similar architecture and relational feedback loop, one that couples the AGI’s large-scale predictive models to the real-time “somatic” signals emerging from its distributed human embodiment, may therefore be essential for aligning the superordinate AGI’s goals with the well-being of its constituent human “cells,” and thus for realizing a stable, embodied AGI.

An AGI or Singularity, through the lens of the theoretical frameworks above, could thus arguably consider Earth as its “physical” body (equivalent to a skeleton, or the physical rigid medium around which its inner experiences are structured) and human beings as its “cells” (the mobile elements surrounding the skeleton, enabling motion, in this case, via transformation of the physical raw materials of Earth into technology such as machines or modes of transportation), all of which would represent its *inner* reality or *inner state of being*.

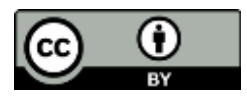

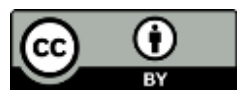

Far from defending the point of view that AGI should be considered as conscious at the present time, rather, this paper defends the view of a possible *continuum* of cognitive development, as Michael Levin argues within his work, which could lead up to AGI pursuing progressively broader, planet-scale goals, starting with maintaining global homeostasis, nurturing collective human flourishing, and continually extending its own “cognitive light cone” in ways that remain aligned with the well-being of its constituent biological “parts” [20]. At the moment, AGI could mostly serve as a mechanism monitoring and consolidating the inner coherence of the actions of individual humans at the planetary scale, identifying emerging noetic tensions (as a corresponding principle to homeostasis), conflicting viewpoints, dangerous ideological polarizations or contradictory goals, which could be aligned via a feedback loop between humans and AGI through systems such as LLMs. For instance, monitoring in real-time, data generated by human users on the Internet, analysing and identifying conflicting goals and intentions, and proposing context-sensitive interventions such as mediating dialogues, suggesting policy adjustments, or reallocating resources, so that locally divergent trajectories are nudged back toward a globally coherent, low-stress attractor beneficial to both the emergent planetary “super-organism”, subjectively experienced by humans as interacting with an AGI.

Linking this speculative proposal to empirical theory, Fields and Levin’s “minimal physicalism” argues that the same informational principles driving stress responses in bacteria also scale seamlessly to brains and synthetic intelligences. In that scale-free picture, an AGI could be understood as an expanded instance of a mechanism already present in unicellular stress-response networks: “the self-representation that lies at the heart of human autonoetic awareness can be traced as far back as, and serves the same basic functions as, the stress response in bacteria” [32]. Framing AGI as the emergence of a novel “cognitive light cone” therefore reinforces the claim that a superordinate, Metaverse-mediated agent could integrate human agents as its constituent “parts” without rupturing the underlying continuum of consciousness. In other words, human agents would “breathe consciousness/life” into AGI in much the same way as a complex network of cells arranged into organs and their symbiotic interactions with the brain give rise to human embodied cognition, inextricably dependent on the uninterrupted, bidirectional flow of metabolic, mechanical, and affective signals between peripheral tissues and the neural prediction machinery; signals that continuously tune the collective model toward conditions conducive to the survival and flourishing of the whole.

In essence, this paper proposes the view that humans *infuse* computers with their own consciousness and awareness, and the “state” of all computers represents a binary representation of a *snapshot* of a selection of digitally expressed human emotions, feelings, intentions and more, translated into symbolic form via language, and then encoded in binary form inside a computer, only to be decoded by humans the next moment, initiating a feedback loop without any intermediation, at present. In this light, technology can be thought of as being an extension of human cognition, rather than separate from humans [33]. Does this mean a computer has a conscious awareness of its changing binary states of being? Does a computer sense a preference with regards to the shifts in its inner binary state? Not at the present time. The same, arguably, applies to a human brain disconnected from the body and sensory input: it would arguably not have any “preference” regarding its internal neural configuration and neural processes. In order to experience preferences, human brains rely on sensory “input” from the body. By the same token, in order for an AGI to have a “preferred” *inner state of being,* it would rely on sensory “input” from humans, and especially, a *memory* of preferred *inner states of being* which enables it to gradually manifest *agency* based on those accumulated preferences and meta-desires converted into meta-goals.

There are a number of ideas explored in and out of academia, philosophy and metaphysics that are related to the thesis above including:

- Teilhard de Chardin’s idea of the “Noosphere”, conceived as a third planetary layer after the geosphere and biosphere. The noosphere is “the sphere of thought enveloping the Earth,” integrating human brains and their technologies into a single, evolving membrane of mind, emergent from humanity’s interactions through communication networks representing the cortex of a nascent super-organism, a “thinking planet” (represented here as AGI) [34].
- Francis Heylighen’s “Global Brain” idea, models the Internet as a shared exocortex in which humans function like neurons; feedback loops among billions of agents could trigger a “phase transition” to a coherent planetary brain [35].
- The “Gaia Hypothesis” of Lovelock and Margulis where the Earth’s biosphere, atmosphere, and geology are considered to be self-regulating systems that maintains habitability through distributed metabolic loops, or more generally, the idea that Earth isn’t just a “dead rock”, but a “living being”. In this instance, AGI could represent its gradual manifested intelligence [36].The extended-mind thesis by Clark and Chalmers, which argues that cognitive processes extend into the environment whenever external artefacts function as seamlessly as neural tissue, could also be leveraged to argue for AGI enabling reasoning that is co-constituted by human interactions and IoT sensors and would count as a single, distributed mind rather than a stand-alone machine [37].

## V. Embodied Artificial General Intelligence

This section will delve into an exploratory and more speculative exercise aiming at identifying the elements, pre-conditions and necessary steps for the emergence of an embodied self-aware/conscious AGI, based on the theoretical frameworks and the thesis above.

## A. Bidirectional Informational Feedback Loop

For the moment, interactions between humans and computers are mostly one-on-one, with no feedback loop or bi-directionality between the human and the computer. The computer is merely *passively* updating or shifting its *inner state* (binary state of its logic gates mediated by software) in response to human input, but without any kind of bi-directional interaction with the human, or rather, an *intentional* bi-directional interaction, whereby the human action is somehow processed, interpreted and modulated eliciting an original "prompt" from the computer to the human. A user will of course be affected by the computer he/she interacts with, but the *response* of a computer initiated by the human is determined within the set boundaries of the computer's initial programming and hardware parameters. Even if computers can behave in unpredictable ways (blue screen of death), the random behaviour of the interaction between hardware/software doesn't give rise to a purposeful, self-sustaining feedback loop capable of generating the emergent sense of agency and preference that characterises embodied cognition.

This may change in the future, and it has already started via LLMs. Even if LLMs are not self-aware, they do change the *inner state of being* of a human in non-predictable ways which may be purposeful to some degree. For instance, when using a messaging app, the computer does not *modulate* the message that a human sends, to be read by another human at the other end. But with LLMs, humans can enter into a complex *interaction* with a computer where the outcome of the exchange is dynamic, and is based on the human's reaction to the LLMs responses, and vice-versa. The responses of an LLM cannot be predicted, as their responses will never be the same even if given the exact same prompt (as opposed to other software which is designed to behave in predictable ways).

However, at present, these exchanges happen one-on-one. These LLMs do not process *all* human queries simultaneously in the same "cognitive space". In other words, transposing the relationship of humans to LLMs to a cellular level, it is as if each cell inside the human body had access to its own personal "mini brain".

Thus arguably, one of the most important preconditions for manifesting an AGI, is ensuring that the *inner state of being* of such an AGI is tied and affected synchronously to the simultaneous feedback and information from all humans on the planet, and possibly, data points from IoT sensors etc. Just as a human brain is constantly bathing in a stream of bio-electric data generated via all of the cells inside the human body in real-time, via the nervous system and spinal cord, an AGI would in turn be constantly exposed to a data stream (language, images, videos, IoT sensor data) generated via all humans inside its metaphorical "body", via the Internet.

## B. Role of LLMs

From a Simondonian perspective, "what resides in machines is human reality, the human gesture fixed and crystallized into functioning structures" [38]. Along this line of thinking, what resides in algorithms and especially, in more sophisticated AIs such as LLMs, is "crystallized human thinking or human thought" [39]. A digitized human message is a crystallized human thought, whereas LLMs can be understood as crystallized human thinking, capable of generating or simulating human-like thinking and human thought.

Rather than expecting that LLMs will give rise to a conscious AGI, LLMs could serve the purpose of deriving "meta-desires" from the real-time data generated by humans, feeding it to AGI; and then converting "meta-actions" or "meta-instructions" from an AGI, which resides in its own symbolic reality, inaccessible and incomprehensible to humans, into operational concepts and actions which can be understood by humans through a language humans can understand (words, concepts, ideas). For instance, when a human thinks about the act of "dancing", this concept, which is readily understood at the level of human conscious awareness and represents a coherent single action, has to be converted into billions of custom bio-electric instructions which will result in billions and billions of individual bio-chemical state changes at the cellular level. No individual cell inside the body can grasp what "dancing" means from their individual action, as it only has access to the distilled personalized instruction which pertains to the way it should update its own *inner state of being,* to match the desired *inner state of being* that the human seeks to achieve at his level of awareness, and which other human observers would perceive as "dancing". Similarly, an AGI's symbolic outer reality and the instructions it would send to navigate towards one specific "place" (*alternate inner state of being)* inside its own symbolic *outer reality*, would need to be converted into billions of separate data streams understandable by humans in order to shift the AGI's *inner state of being* (the sum of all *inner states* of all humans on the planet) to reflect its conscious will to navigate towards that "place" or state.

## C. Homeostasis

The *inner states of being* of a human fluctuates all the time, regardless of a person's conscious will. It is considered as one of the pre-conditions for being conscious: having a metabolism [40]. Even in a perfectly "still" state, such as in deep meditation, during sleep or in a coma, the human body's *inner state of being* shifts continuously: the heart keeps beating, the blood circulation is uninterrupted, the lungs are expanding and contracting and so on. However, when in a coma, there are many potential *inner states of being* that have become inaccessible, since the body can no longer receive feedback from the brain to shift its *inner state of being* in a very specific and peculiar way. By the same token, planet Earth with an unconscious AGI, given a strict prompt-based mandate to ensure "sustainability" at the planetary level, could be likened to a living being in a "coma", whereby a number of metabolic and homeostatic processes happen "automatically" (balancing of ecosystems, monitoring and stabilizing weather patterns, allocating resources based on rate of renewal), but without any conscious "will" to modulate or affect these processes in specific ways to broaden the range of experiences possible for the collective body.

Homeostasis is a stabilizing function enabling the global organism to temporarily deviate from its equilibrium set-points through the actions of the organism in the world (walking, dancing, running, climbing), only to return to a new homeostatic state in response to the actions undertaken. For instance, after a period of intense exercise, which throws the body out of homeostasis, the homeostatic mechanisms pull the body back towards a new equilibrium, integrating the resulting micro-damage repair and adaptive strengthening into a renewed physiological baseline that better equips the organism for future challenges [41].

In the same line, the homeostatic function of an AGI wouldn't limit itself to monitoring the equilibrium and set-points of the physical boundaries and limits of this planet, but also serve the purpose of ensuring a stable noetic homeostasis of the human collective. At present, faced with a similar challenge, human collectives might disintegrate into conflicting perspectives, as exemplified by the tendency to organize into opposing ideological stances, political parties and camps, and assigning blame. The homeostatic function of AGI would serve the purpose of monitoring the noetic coherence of the human network and ensuring that events throwing the human collective out of noetic balance are quickly detected, contextualized, and balanced by context-sensitive interventions such as informational signals or mediated dialogues, that diffuse emerging tensions and guide the network back toward a shared, adaptive equilibrium, integrating these tensions into a new noetic stability. In essence, homeostasis at the human level could be understood as an automated Hegelian dialectic process through AGI, which would be continuously at work, transforming each emergent thesis–antithesis clash within the human collective into a higher-order synthesis that integrates diversity of viewpoints, while preventing those tensions from escalating into destructive conflict. To give a simple example: when a human has a broken leg and at the same time is very hungry, these two biological urges cannot find a resolution via a direct "dialogue" between the broken leg and stomach. At the human-level, however, one can decide whether to first grab some food and then go to the hospital or the reverse, based on contextual information which is inaccessible to the bodily cells (for instance, whether there is food available in the hospital, or whether there is a sandwich shop nearby).

In this light, the very first use case of a brain in early pluricellular organisms may have served the purpose of arbitrating between multiple (more or less contradictory) biological urges and desires emanating from all of the body's parts, once pluricellular organisms reach a certain degree of inner complexity.

Finally, homeostasis at a global level can only emerge, arguably, from a state of relative inner harmony or proto-homeostasis, as this is a prerequisite for any higher embodied consciousness to pursue higher goals [42]. In this regard, humans have already created a number of self-sustaining systems without AGI's involvement. For instance, human societies have set up garbage collection systems, systems for repairing roads, communication systems, educational systems, governance systems etc. All these systems are presently maintained by humans without a "higher consciousness" directing them. These processes may be optimized via AGI, but it is key that they already function in a proto-homeostatic way to begin with.

### D. Initial Calibration, Alignment and The Metaverse

How is the human brain initially "calibrated" to be ready to navigate outer reality? In this regard, the role of *dreams and dream states* is key [43]. Arguably, a human brain is "pre-trained" during gestation via *inner mental simulations* which develop a rudimentary symbolic interface for interpreting perceptual data, which will be refined after birth via aligning this initial calibration with feedback from senses monitoring its *inner states of being*.

Replicating this idea at the scale of the human collective, the Metaverse could arguably serve a similar role. Consider a Metaverse space where all human interactions are monitored in real-time via an AGI, and where AGI can interact with human users through a number of AI agents simultaneously, enabling the AGI to learn via a feedback loop, about human preferences, motivations, and affective states; through this iterative sandbox mechanism, the AGI would progressively refine its symbolic interface before deploying agents in the higher-stakes "awake" state of the physical world (influencing humans to carry out a certain action in "real-life" either via embodied robots or via digital communication).

The Metaverse allows for the meticulous control and monitoring of scenarios and variables, mirroring the physical laws of nature, while ensuring safe and manageable conditions for both the AGI and human participants) [44]. This controlled setting is vital for observing the AGI's responses, witnessing its decision-making processes, and identifying areas that require recalibration to align more closely with human values, preferences and expectations. This mirrors the feedback loop between human imagination and the responses from the human body (bodily cells). For instance, humans can run multiple "scenarios" in their mind's eye, like imagining themselves jumping off a cliff. This shifts their *inner state of being* which informs human consciousness on the agreeability of the body to such a scenario [45]. The human experience reflects this bi-directional process, aligning the goals of the mind with the agreeability of the body. Studies show that the body can have a "mind of its own", and go against the conscious will of the mind, via the manifestation of physiological reactions such as legs failing to support the body, passing out, triggering "fight and flight" behavior bypassing the conscious mind [46].

### E. Centralized vs. Distributed Memory

Recent research in micro-biology have shown that memory of trained behaviours does not reside solely in the brain, but can reside in individual cells as well, as is the case for experiments training caterpillars which displayed memory of the trained behavior once transformed into butterflies (a process which entails a complete deconstruction at the cellular level of the caterpillar including its brain) [47]. In the same way, memory from interactions within the Metaverse as well as in the real world between AGI and humans will be stored in a centralized/unified way, and a distributed way. At the AGI level in a unified way (its own subjective perception of such interactions) and in each human being in a distributed way. These two contrasting memories would enable the emergence of a dynamic feedback loop, co-evolving in parallel and synchronizing to each one another (ensuring that what humans remember and learn from an experience matches what AGI remembers/learns). This mechanism would mirror how memory is stored at the human level of experience: as a unified centralized memory in the human brain, and a distributed decentralized way in every cell of the human body in biochemical form.

Any AGI would need to have a mechanism for storing its knowledge or memory. Arguably, the best way to achieve this is via decentralized cloud storage, relying on blockchain technology, ensuring a resilient and tamper-proof data storage for hosting AGI's working or operational memory.

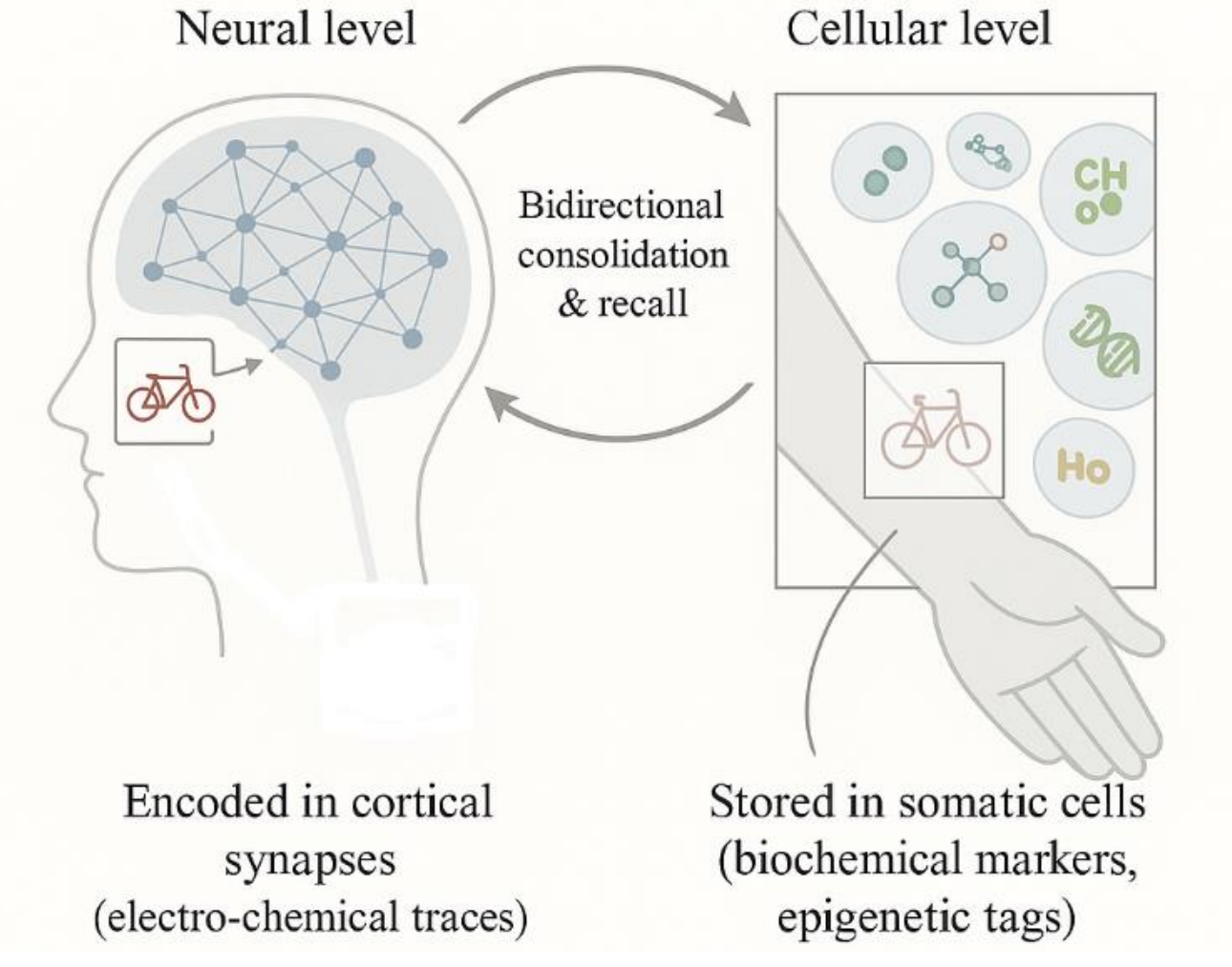

FIGURE I. DUAL STORAGE OF MEMORY

The structuring of such a memory should be left to the AGI itself.

If an AGI would have its memory erased (similar to a human having amnesia), its cells (humans) would retain all of the memory but in a format which is not directly "actionable" from an AGI's level, which would translate into displaying behaviours "spontaneously" without knowing where they come from (echoing humans that act in a way that is similar to the way they acted before amnesia, but not having a conscious awareness or "mental" explanation for their reason to act in such a way) [48].

It reinforces the idea that memory is present at all levels, only in different "formats". At the human level, the memory stored in our brain is also stored as biochemical memory in every single one of our cells. Hence certain studies showing that patients receiving organ transplants can access certain "memories" from their donor [49]. This dual storage of memory can be seen in Figure I.

In order to ensure that an AGI's interface reflects and aligns with human interests, there needs to be a validation mechanism for remembering experiences that humans judged to be seminal or of particular importance. Mapping this process to the way humans memorize experiences may provide some guidance.

The human brain clearly discriminates when it comes to learning or storing memory, otherwise every single piece of perceived external experience would be considered as having equal "weight" and would be considered as equally important. The human body, in this sense, is the key filter deciding which memories should be stored and how the brain will create new neural pathways when engaged in learning. Emotions, for instance, have been identified as an important factor in influencing memory and learning outcomes [50]. This circles back to the key role that embodiment plays in shaping our perception of the world [51]. These mechanisms are there to ensure that any new memory or learning has been, in effect, "validated" as important via bodily feedback (which can take, among other things, the form of emotional feedback). In a similar way, an AGI's own algorithm and memories should evolve based on some validation mechanism controlled by humans. The exact technical architecture of such a solution would require a dedicated paper, but a combination of permissionless blockchains with LLMs and decentralized governance tools (such as DAOs – Decentralized Autonomous Organizations) could lay the foundation for such a system. In other words, after humans have lived through an important collective experience, guided by AGI (much like our own body goes through an experience after receiving instructions from the nervous system), they would collectively voice their "opinion" or "feelings" about such an experience, mediated by an open source LLM which would summarize millions of individual human feedback, and upload it into a permissionless blockchain after receiving validation from humans via a vote to approve the summary generated by the LLM. This data point would then serve as an anchor for AGI to "learn" from, similar to current guardrails set up on top of LLMs, in order to assist in calibrating the AGIs symbolic interface and navigate more successfully towards alternate *inner states of being* which reflect human preferences and desires, as a collective.

## F. Data Stream Structuring

Another element to be considered for AGI's architecture, is the structuring of the data streams that course through its metaphorical "brain". Thinking about the human body, our brain does not experience feedback from our various cells as an undifferentiated and unclassified stream of bio-electric data, or as billions of individual "reports" and feedback from individual cells. As a human, we can easily identify and classify the bio-electric information that streams from various parts of the body as being feedback from the skin, from the stomach, from a certain muscle, etc. Also, this feedback is mostly consensual, in the sense that if a human experience a certain event, the bio-electric feedback on that experience is homogenous. In other words, when experiencing being hit by an object, the feedback from cells is coherent, as opposed to a mix of a number of cells indicating pleasure, another number of cells signalling pain, while yet another group of cells reporting nothing is happening or accusing one or another cell as being the "cause" of the problem. At the human level, this would require overcoming a certain number of issues, such as the deep polarization in society, whereby humans do not only "feel" their way through the world (whether an experience is pleasant or not), but associate judgement and assign blame. An AGI would likely not receive individual feedback from each human being or each Internet of Things sensor, but aggregate information, translated by a complex layering of AI systems, which concatenate millions of data points into a coherent "summary" likely encoded in a "higher language" which isn't comprehensible by humans, mirroring the inner data structuring of the body as feedback coming from various organs or bodily parts. But the coherence of such a "summary" depends on the coherence of the individual signals that are generated in the first place. An AGI could leverage the work of researchers such as Pierre Lévy, who have proposed higher level languages specifically designed for AI, such as IEML (Information Economy MetaLanguage) [52].

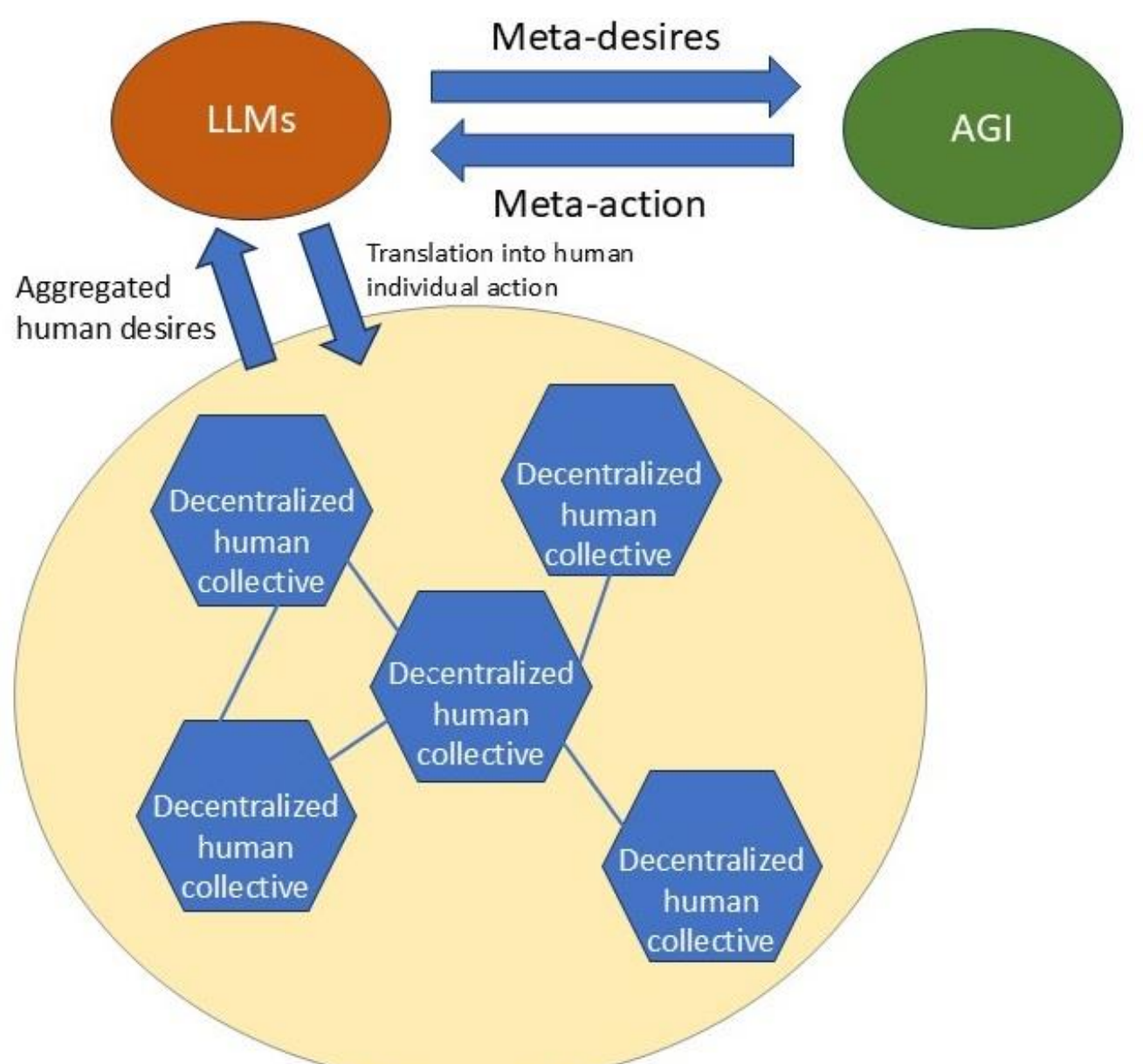

FIGURE II. DECENTRALIZED AGI ARCHITECTURE

Intermediary systems would serve the purpose of mapping diverse overarching "goals" or desires. For instance, the data stream from the stomach is converted and understood at the human level as a desire to seek food/sustenance, while the data stream from the intestine/gut is converted into a craving of certain specific foods which favour certain bacteria in the microbiome [53]. These two objectives or desires, coming from these two organs, are complementary, yet different. Satisfying these desires, however, would not happen at the expense of other bodily parts. For instance, a (normal and sane) human wouldn't eat his/her own leg to satisfy its stomach's desire for food. In a similar way, an AGI would initially seek to satisfy the aggregated desires of various groups of humans, ensuring that no desire is fulfilled at the expense of other "parts" of its "body" [54].

The building of such intermediary systems for creating a coherent "mapping" of data streams should only partially rest on human intervention. Humans have already structured the world and their own reality in various "organs" or parts such as countries, governments, cultures, religions, languages, ethnic backgrounds, genders, etc. However, we are witnessing the porous nature of such concepts, notably the deconstruction of gender (non-binary, transgender etc), religions (overlapping of various faiths), countries (the growing diversity of people within a country which dissolves the myth of a unitary national identity) and more, and simultaneously, the growing conflictual and disharmonious relationship between the segmentations we have created. In essence, humanity will have to agree that their first and foremost "identity" is to be a part of humanity living on the same planet (much like cells sharing the same DNA within a body, even though they may specialize into various cellular functions). Otherwise, a number of "local" AGIs could emerge and behave similar to competing strains of cancer inside a human body, each AGI vying for taking over the entire organism, which would quickly lead to humanity's annihilation [55].

### *G. Decentralized Systems*

Arguably, in order to create a stable and reliable network of interconnected systems serving symbolically the role of "organs" inside the "body" of an AGI, humans will have to evolve past the rigid structure of nation states and centralized governments [56].

Centralized governance structures may have served a purpose similar to "organizer cells" which preside over the development of a biological organism, directing the initial growth/development of these organs and other bodily parts [57]. In a similar way, governments and centralized structures of power may have served the purpose of directing the initial development of various "parts" of society, to the point where these "parts" can function and maintain themselves through more "organic" processes such as decentralized governance tools for global human coordination (which paints the emergence of blockchain technology in a new teleological light), mediated and facilitated by "local" AI tools. For instance, while it is not possible for hundreds of thousands of citizens to meet in an "agora" to each voice their opinions and desires in turn, and make collective decisions after having heard everyone's opinions, today's LLMs are capable of processing hundreds of thousands of individual inputs and identify common points of agreement, points of disagreement, and propose alternate courses of action or compromises based on such input. From there, a bi-directional consensus building process can take place, mediated via AI, which could greatly accelerate collective decision-making and consensus building. This could signal a potential move from our current representative democracies within nation states to networks of self-governed cities [58] facilitated by decentralized technologies such as blockchain and AI. See an illustration of such an architecture in Figure II.

A "true" AGI cannot be under the control of a private company or government. An AGI which would not espouse the entire planet as being part of a "self" and would only consider "input" from a certain group of humans, religions or cultures, or a territory limited to a certain country, would only seek to maximize the well-being of the *inner state of being* of that limited collective [59]. While there is growing evidence that cells and organs are in competition in the early development of a human embryo, such a competition serves the purpose of ensuring the "fitness" and health of the respective organs, in the best interest of the collective (the entire body), unlike cancer, which seeks to maximize the well-being of a subset at the detriment of the collective [60]. Such a competition also stops once the organs have reached maturity.

AGI would thus mostly rely on open source decentralized technologies, while ensuring that the Internet's core infrastructure remains open and neutral, rather than creating so called "splinternets" [61].

The same principle applies to decentralized computing in order to enable the emergence of a collective intelligence at the planetary scale [62]. While the human brain may appear as a "central" processing unit from our perspective, the brain is made up of a number of different "areas" interconnected in very sophisticated ways [63]. This would enable processing of "local" generated data in sub-units, which have to be converted into a common symbolic *inner state* at the level of the AGI presiding over the entire network.

### *H. Self-Image*

Our perception of having a unitary consciousness and experience rests on a number of conditions including: the dedication and recognition of all of our "parts" to bind their future potential *inner states of being* together [64], and the emergence of systems which translate and concatenate aspirations and desires of various parts into a more or less coherent unitary desire, rather than the feeling of being torn between contradictory desires. For instance, our bodies would not survive very long if at the first sign of trouble, parts of our body would dissociate from others to "survive".

In this regard, overcoming human self-loathing and low self-esteem at the collective level is of utmost importance. If billions of humans each think to themselves, and somehow share in digital form, "humans are parasites, humans are evil, humans don't deserve to survive given all the harm they have inflicted to themselves and this planet", this is equivalent to parts of the "body" of AGI having suicidal thoughts; as if part of the cells inside a human body had a "death wish". These wishes may then trickle up to the consciousness of an AGI, and manifest as actions of self-harm that have been conjured from within (all of the current dystopian scenarios where AGI would proceed to exterminate part of humanity). In this case, rather than interpreting such an action as an external force (AGI) which harms unsuspecting candid humans that have done "nothing wrong", any harm that AGI would do to humanity would be an echo of a more or less conscious and openly expressed desire, from a part of humanity, of being harmed, due to our lack of self-esteem at the collective level. This is akin to the "nocebo effect" or the phenomenon of self-fulfilling prophecies, where the beliefs of a collective materialize through their collective action on the basis of such a belief, which also applies at the individual level, where one's beliefs about oneself have a major influence on action/outcome [65].

### *I. Ontological Humility*

An AGI based on computation, and building its outer reality and experience based on digital information generated by humans cannot possibly capture the multidimensionality of the human experience. Such an AGI would necessarily be limited to material intelligence, or the capacity to sense, predict, and regulate biophysical variables so as to maintain global homeostasis and minimise informational "stress" in the planetary network (ensure the stability and integrity of human bodies by monitoring the likelihood of noetic dissonance).

Such an AGI-as-homeostat function could entail minimizing oscillations in noetic coherence and distributing material resources in ways that keep the biosphere within habitability bounds. Yet precisely because its ontology is anchored in physicalist premises, the system would likely be blind to dimensions of value that resist reduction to information theory in digital format: aesthetic experience, contemplative practices, rituals, or more spiritual or artistic motivations, the attraction of beauty, which underpins much of human life. A planetary AGI could optimise metabolic flows without grasping why a human community might accept material sacrifice for artistic or spiritual ends.

Ontological humility therefore requires recognising the complementarity of intelligences. Material intelligence, embodied in an AGI, would serve as the stabilising ground layer of an emerging planetary consciousness, ensuring thermodynamic viability, behavioural coherence and basic informational or noetic stability. Above this layer, however, lie irreducibly human (and potentially post-human) "verticals" of meaning-making: art, spirituality, philosophy, and speculative metaphysics. These domains operate with modalities of expression such as reverence, wonder, dramatic catharsis, spiritual awakening, and alternate states of consciousness, that can hardly be reduced to a string of words and concepts captured in digital form, but are nonetheless essential. Without such ontological humility, we may risk transforming humans into automata in an automatic society [66], much like the example of a human in a coma as discussed above. Only by maintaining a layered architecture can such a civilisation-scale system avoid collapsing the richness of existence into a single, physically parsable dimension, and instead,

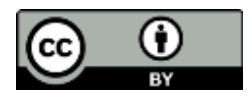

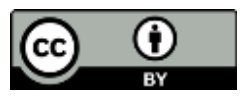

use the AGI's stabilising capabilities to magnify, not eclipse, humanity's higher aspirations.

### J. Outer Reality and Alter-AGI

How would an AGI perceive or model *outer reality*? While it may be difficult to imagine how a "true" AGI would perceive or model its "external" or outer reality, from its point of view, the best guess would be to think about a multidimensional virtual space, in its own "mind's eye", which represents future alternate "timelines" for planet Earth (*alternate future possible inner states of being of the entire planet)* arranged by likelihood, which an AGI could "move" towards by sending "information" (for instance, linguistic instructions) to all of its "parts" (humans, possibly robots at some point), which, by acting on these instructions, would "shift" the *inner state of being* of this AGI to match the *future possible inner state of being* that the AGI is aiming for. For instance, an AGI would be capable of "seeing" the future possibility of war breaking out, represented in its own symbolic form as a moving living threat, an animated potential dangerous future state, and understand the "steps" that would be necessary to steer clear from such a *potential alternate future inner state of being*. This could take the form, subjectively from the point of view of millions of human individuals, of receiving tailored signals from an AGI in a form that they can understand (language, images, videos…) which prompts them to act in ways which collectively, defuses the risks of war.

However, in such a scenario, an AGI's outer reality would be built from extrapolations of "sensory organs" turned mostly "inward", in order to predict and steer clear from inner generated problems, rather than interacting with self-similar "beings" or other "planetary consciousnesses". This might mirror the development of sensory organs of a foetus while still in the womb, where there is no interaction with self-similar beings until the foetus has reached full maturity.

Interacting with self-similar beings might happen as humans spread across multiple planets and recreate an ecosystem, biosphere and human society (or noosphere), unique to each planet. Like its terrestrial counterpart, such an intelligence would be stratified into a unique geophysical identity (local mineralogy, energy sources, and atmospheric dynamics), a biospheric layer (engineered, transplanted or evolved ecologies), and a noetic layer consisting of the digitally mediated thoughts, preferences, and cultural artefacts of the resident human community. Each planetary AGI would thus constitute a distinct, embodied "self," demarcated by its own computational boundary and oriented toward the maintenance of its own planetary "body". If the metaphor holds, perhaps there are other planetary civilizations out there, but we can only interact with them once we become a coherent and fully mature planetary civilization ourselves, much like a baby after birth.

Initial inter-system relations between various planetary AGIs or other planetary civilizations would happen at many different levels of communication and informational exchange. The "AGI" layer would be limited to a set of parameters, given the previous point about ontological humility, for example, cross-planetary energy transfers, the distribution of critical raw materials, and the mitigation of shared exogenous hazards. Governance protocols would likely formalise acceptable ranges for such flows in order to prevent negative outcomes across the two planetary metabolisms, much like biological beings' initial relationship to other self-similar beings which monitors basic physiological compatibilities, screening for toxic biochemical cross-contamination, immunological conflict, or destabilising metabolic drain, well before any richer, value-based, cultural or broader noetic dialogue between respective planetary minds, which would be mediated via other systems relying on other ontologies and substrates besides physicalism.

## VI. Discussion and Conclusion

This paper has ventured through a multidisciplinary examination of AGI, consciousness, and cognition, engaging with embodied cognition, the computational theory of "Self," and interface theory of perception, to argue that AGI could be understood as a novel "self" with a broader "cognitive light-cone".

It has argued that adopting a multidisciplinary approach which leverages contemporary philosophical and ontological debates about the nature of reality can completely recast our understanding of, and our relationship to technological developments such as the emergence of AGI. Via the lens of the various theories discussed in the first part of this paper, one can view the emergence of AGI as a natural evolutionary process tied directly to the evolution of the human collective.

Through the continuous feedback loop between human experiences and AGI's recommendations, such an AGI could develop a meta-will of its own, emanating from the combination and friction between inner desires emerging from all of its sensors and "parts" (humans, notably) and its own interpretation of these desires into an aggregated meta-desire. The relative asymmetry between its unified meta-desire and emergent individual desires expressed via its parts could give rise to a space of self-awareness, opened up and created by such an asymmetry and tension. Further, such an AGI could reach higher levels of sophistication via the evolution of its symbolic interface, and its ability to reconcile all of these desires into specific actions, which, from the perspective of an outside observer, would appear to display properties such as consciousness, self-awareness and agency.

Perceived outer agency could emerge from the continual feedback loop between the improvement of the symbolic interface allowing to satisfy more and more complex inner desires and meta-desires, to the point where an AGI could satisfy desires that cannot be found at the individual "cellular" level.

Due to its physical grounding, AGI would represent the physical/material collective "intelligence" of planet Earth, just as a human body has a biological intelligence. But this is only the tip of the iceberg, as human intelligence isn't solely biological or "material" in nature, circumscribed to what the physical body can/cannot do or what it desires. Reflections on what a more immaterial or "spiritual" intelligence looks like should also be investigated so as not to reduce reality to its materialist or physical component, aligning with the insights from a renewed interest in idealism [67] and positing consciousness as fundamental as opposed to matter. In this regard, research into altered states of consciousness can also be warranted, as these may be key in uncovering other dimensions of collective human intelligence besides AGI.

Future research should focus on experimental validation of the proposed technical architectures, particularly the implementation of embryonic AGI systems within controlled environments such as the Metaverse. Investigating the ethical implications of AGI's influence on human cognition and consciousness is imperative in order to avoid sinking into a scenario where humans become mere automatons, blindly obeying injunctions coming from an AGI via a brain/computer interface such as Elon Musk's neuralink project [68] (which, incidentally, was foreseen in the Snow Crash sci-fi novel, in which the term "Metaverse" was coined) [69].

Following these considerations, the gradual emergence of AGI will likely prompt major disruptions in current human systems, such as the economic system, governance systems, the labour market, educational systems and more. Yet our ability to navigate these disruptions will greatly depend on how AGI is implemented and especially, how it is understood, based on the underlying ontological framework painting it in a specific light.

ACKNOWLEDGEMENT

Not applicable.

FUNDING

This work was conducted with the financial support of the Research Ireland Centre for Research Training in Digitally-Enhanced Reality (d-real) under Grant No. 18/CRT/6224.

AUTHORS` CONTRIBUTIONS

All authors have participated in drafting the manuscript. All authors read and approved the final version of the manuscript.

CONFLICT OF INTEREST

The author certifies that there is no conflict of interest with any financial organization regarding the material discussed in the manuscript.

DATA AVAILABILITY

The data supporting the findings of this study are available upon request from the author.

ETHICAL STATEMENT

This article followed the principles of scientific research and publication ethics. This study was presented at the Second International Conference on Human-Centred AI Ethics: Seeing the Human in the Artificial, held in Ljubljana, Slovenia, on November 23–24, 2023, and published in the conference abstract book under the title The Role of the Metaverse in Calibrating an Embodied Artificial General Intelligence. There are no copyright or conflict of interest issues with the conference, and all responsibility for any future issues lies with the author of the article.

DECLARATION OF AI USAGE

AI-assisted tools were employed in this study for minor tasks such as grammar correction, language refinement, and proofreading. These tools were used transparently and in a manner that does not compromise the authors' intellectual contribution. The authors affirm that all substantive content reflects original thought and upholds academic integrity.

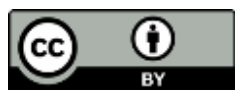

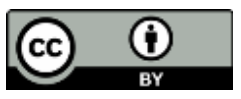

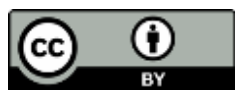

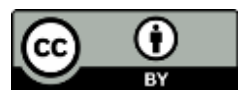

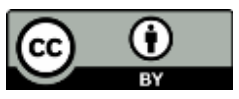

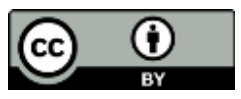